\documentclass[runningheads]{llncs}

\usepackage{eccv}

\usepackage{eccvabbrv}
\usepackage{hyperref}
\usepackage{multirow}
\usepackage{pifont}
\usepackage{booktabs}
\usepackage[accsupp]{axessibility}  

\usepackage{hyperref}

\usepackage{orcidlink}

\begin{document}

\title{Exo2EgoSyn: Unlocking Foundation Video Generation Models for \\ Exocentric-to-Egocentric Video Synthesis} 

\titlerunning{Exo2EgoSyn}

\author{Mohammad Mahdi\inst{1} \quad Yuqian Fu\inst{1} \quad Nedko Savov\inst{1} \quad Jiancheng Pan\inst{1} \quad \quad
Danda Pani Paudel\inst{1}  \quad Luc Van Gool\inst{1}}

\authorrunning{Mahdi et al.}

\institute{$^1$INSAIT, Sofia University “St. Kliment Ohridski” \\
\texttt{firstname.lastname@insait.ai}}

\maketitle

\begin{abstract}
  {Foundation video generation models, such as WAN2.2 exhibit strong text- and image-conditioned synthesis abilities, but remain constrained to the same-view generation setting. In this work, we introduce Exo2EgoSyn, an adaptation of WAN2.2 that unlocks Exocentric-to-Egocentric (Exo2Ego) cross-view video synthesis. Our framework consists of three key modules. Ego-Exo View Alignment (EgoExo-Align) enforces latent-space alignment between exocentric and egocentric first-frame representations, re-orienting the generative space from the given exo view toward the ego view. Multi-view Exocentric Video Conditioning (MultiExoCon) aggregates multi-view exocentric videos into a unified conditioning signal, extending WAN2.2 beyond its vanilla single-image or text conditioning. Furthermore, Pose-Aware Latent Injection (PoseInj) injects relative exo-to-ego camera pose information into the latent state, guiding geometry-aware synthesis across viewpoints. Together, these modules enable high-fidelity ego-view video generation from third-person observations without retraining from scratch. Experiments on ExoEgo4D validate that Exo2EgoSyn significantly improves Exo2Ego synthesis, paving the way for scalable cross-view video generation with foundation models. The code will be released at \url{https://github.com/insait-institute/Exo2EgoSyn}.}
\end{abstract}

\section{Introduction}
\label{sec:intro}
Exocentric-to-egocentric (Exo2Ego) cross-view video generation aims to synthesize how the world would appear from a first-person viewpoint given a third-person observation of the same scene, with promising applications in robotics, augmented reality, and virtual reality. {This task requires not only transferring the visual perspective, but also inferring fine-grained spatial and temporal cues, such as body dynamics and hand–object interactions, and recovering details that are only partially observable from exocentric views, making it a challenging and largely unsolved problem.}

\begin{figure}[t]
  \centering
 \includegraphics[width=0.5\linewidth]{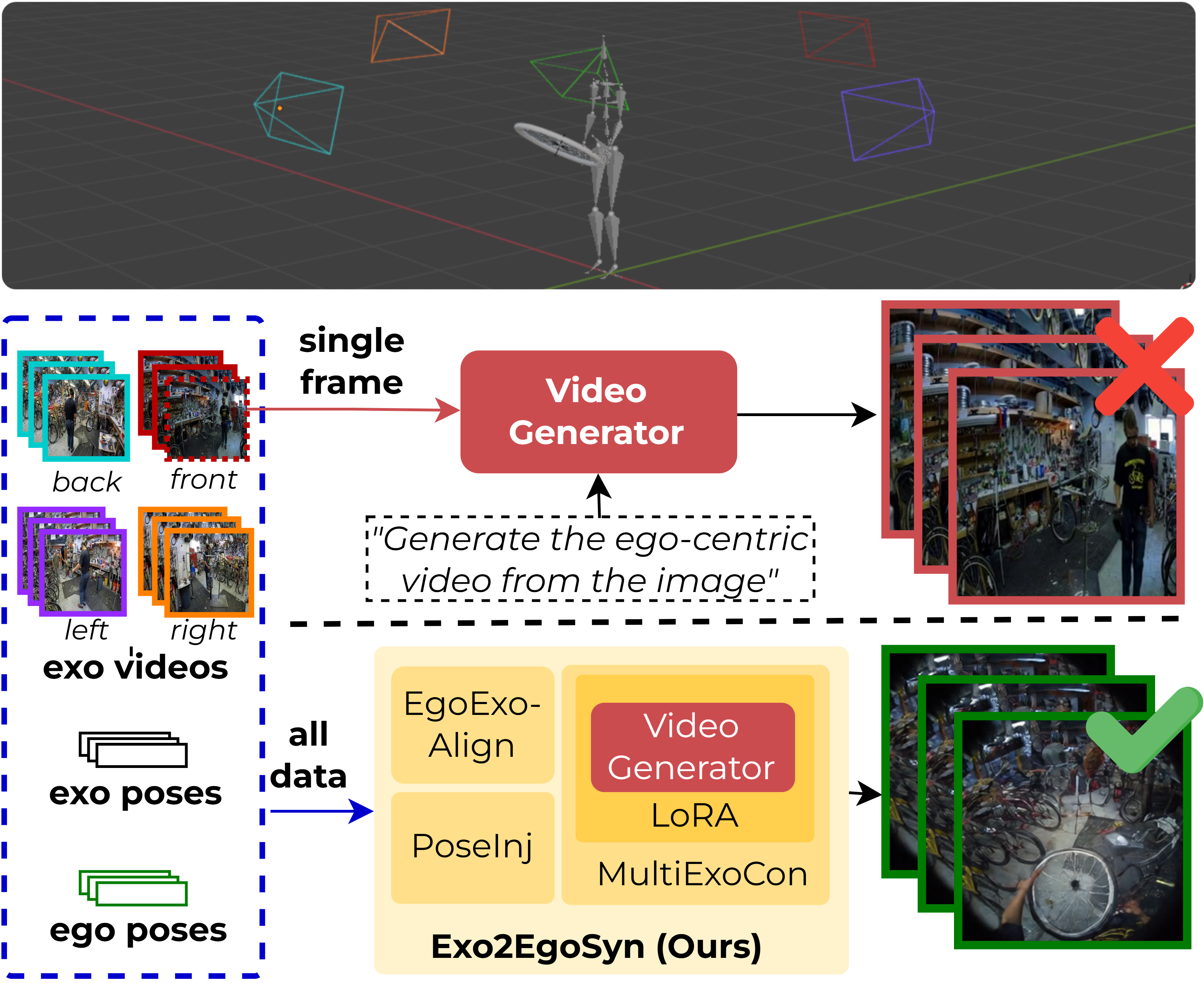}
  \caption{\textbf{Adapting a video generator for Exo2Ego task with Exo2EgoSyn.} Top: Multi-view (4 exo, 1 ego) setup. Middle: The original generator cannot perform cross-view translation. Bottom: Our model Exo2EgoSyn enables Exo2Ego cross-view translation. }
  \label{fig:fig1}
  \vspace{-0.2in}
\end{figure}

{
Few efforts~\cite{xu2025egoexo,luo2024intention,liu2024exocentric,park2025egoworld} have been made to address this task; however, most existing approaches design task-tailored models and train them from scratch on cross-view paired videos, typically from Ego-Exo4D~\cite{grauman2024ego}. The limited training source inevitably restricts their generalization ability. In contrast, a number of emerging foundation video generation models, e.g., WAN2.2~\cite{wan2025wan}, trained on large-scale and diverse video corpora with strong generalization even under zero-shot settings. Yet, these foundation models are inherently designed for single-view generation, leaving their potential adaptation to cross-view scenarios largely unexplored. This gap motivates a central question: \textit{Can we unlock the capability of current foundation video generation models for cross-view scenarios, particularly for the Exo2Ego video generation task?}
}

{
To achieve this, we must address two key obstacles:
\ding{172} \textbf{Strong view bias.} Foundation video generation models are trained on massive datasets consisting of individual videos, which leads to inherent view biases. More specifically, when conditioned on an image, such models tend to reproduce subsequent frames from the \emph{same} viewpoint or camera pose as the conditioning image{, as in Fig.~\ref{fig:fig1}}. This behavior conflicts with the cross-view generation setup, where the target video should begin from a different viewpoint rather than the conditioned one.
\ding{173} \textbf{Limited conditioning support.} Most existing foundation models allow video generation conditioned on text descriptions or a single image, leaving novel conditioning signals, such as multi-view exocentric observations or camera poses, unexplored. Although prior works~\cite{bahmani2025ac3d,bahmani2024vd3d} have explored incorporating new conditioning types, they typically assume access to a model already fine-tuned on its standard conditions (e.g., text), which is not the case in our scenario.
}

{In this work, we start with WAN2.2, one of the most recent, flagship and high-performing video generation models, and use it as an example to tackle the Exo2Ego video generation task by addressing the two aforementioned challenges. Specifically, to tackle \ding{172}, we note that the first frame, which serves as the sole image condition, plays a crucial role in determining the viewpoint or camera space of the generated video. Instead of avoiding this inherent bias, we propose to leverage it to facilitate Exo2Ego translation. To this end, we introduce an Ego–Exo View Alignment (EgoExo-Align) module that enforces alignment between the latent representations of the first frames from the ego- and exocentric videos. Although seemingly straightforward, this design effectively guides the model to shift its generation space from the exocentric to the egocentric view. Furthermore, to fully exploit the information available in the Exo2Ego video generation task, we condition the model on multiple exocentric videos and known camera poses. For this purpose, we propose two additional components: Multi-view Exocentric Video Conditioning (MultiExoCon) and Pose-Aware Latent Injection (PoseInj), both addressing \ding{173}. In particular, MultiExoCon is implemented by replacing the original supported text tokens with video latent tokens from the exocentric videos, while PoseInj is integrated into the network by {first lifting its sparse representation into a dense form and then combining it with the hidden state of {the diffusion} model,} bridging the gap between sparse pose information and dense feature representations. Notably, we focus on the relative pose between the exocentric and egocentric views, which naturally aligns with the objective of the Exo2Ego translation task.}

{
Formally, with all the newly proposed modules integrated in the vanilla WAN2.2, we present a novel framework, termed \textbf{Exo2EgoSyn}, for exocentric-to-egocentric video generation. To the best of our knowledge, this is the first approach that successfully explores and extends a single-view foundation video generation model to cross-view generation tasks, thereby broadening its applicability and generalization capability.
{Extensive experiments on the challenging Ego-Exo4D benchmark show that Exo2EgoSyn consistently advances beyond the proposed baseline, delivering substantial performance improvements in this task.}}



In summary, our contributions are threefold:
{1) We present Exo2EgoSyn, the first framework that extends a single-view foundation video generation model to the exocentric-to-egocentric video generation task.
2) We design three key modules, including EgoExo-Align, MultiExoCon, and PoseInj, enabling effective view translation and multi-video, pose-aware conditioning.
}
{3) Extensive experiments on the Ego-Exo4D benchmark show that our approach outperforms the baseline, delivering clear performance gains.}

\begin{figure*}[t]
  \centering
  \includegraphics[width=\linewidth]{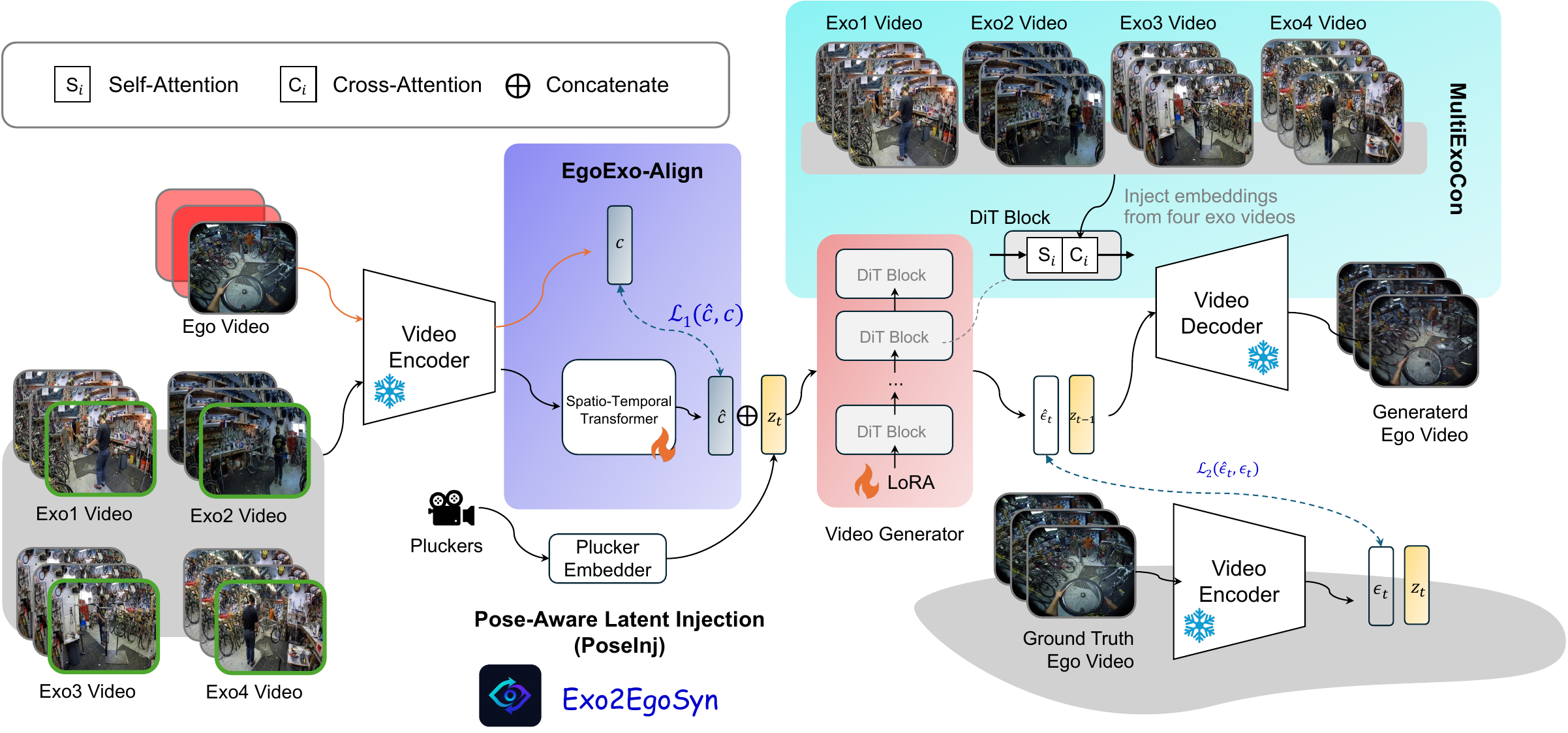}
  \caption{\textbf{Exo2EgoSyn Framework.} We first propose the EgoExo-Align module, which predicts the latent representation of the first frame of the ego video. The MultiExoCon module replaces the original text conditioning in the foundation model with latent tokens derived from exocentric videos, injected into each DiT block of the diffusion model. During the second stage of fine-tuning, we concatenate the Plücker embeddings of the relative camera poses with the hidden states of the diffusion model before passing them to the first DiT block. }
  \label{fig:fig2}
\end{figure*}

\section{Related Works}
\label{sec:relatedworks}

\noindent{\textbf{Diffusion based Video Generation.}}
{Recent advances in diffusion models~\cite{rombach2022high,huang2024learning,liu2025control,liu2025diverse,pan2025earthsynth,blattmann2023align,ho2022imagen,ho2022video,zhang2025show,zhou2022magicvideo,peebles2023scalable,yang2024cogvideox} and large-scale datasets have enabled the rapid development of video generation. Flagship methods such as Tune-A-Video~\cite{wu2023tune}, Video Diffusion Models~\cite{ho2022video}, Stable Video Diffusion~\cite{blattmann2023stable}, AnimateDiff~\cite{guo2023animatediff}, and WAN2.2~\cite{wan2025wan}.  Notably, WAN2.2 stands out with its high performance and strong generalization even in zero-shot settings. However, while these models excel at single-view video generation, their potential for cross-view scenarios, including multi-view exocentric input and exo-to-ego pose conditioning~\cite{kant2024spad,he2025cameractrl,kong2024eschernet,bai2025recammaster,kang2025multi,jiang2025rayzer,li2025cameras,bahmani2024vd3d,bahmani2025ac3d,jiang2025vace}, remains largely unexplored. We aim to close this gap by adapting WAN2.2 for robust Exo2Ego video synthesis.}

\noindent\textbf{Ego-exo Cross-View Learning.}
The release of ego–exo dual-view datasets~\cite{grauman2024ego, huang2024egoexolearn}, notably Ego-Exo4D~\cite{grauman2024ego}, enables new forms of cross-view understanding. Prior work studies ego–exo object correspondence~\cite{fu2025objectrelator, fu2025cross, mur2025mama, pan2025v2sam} and ego–exo VQA~\cite{he2025egoexobench, lee2025towards}. In contrast, we focus on {Ego–Exo Video Generation}~\cite{liu2024exocentric}, a more challenging synthesis task beyond perception and reasoning.
Despite its importance, only a few recent works study ego–exo video generation. Luo et al.~\cite{luo2024put} first predict hand trajectories, then generate egocentric videos from exocentric inputs. EgoWorld~\cite{park2025egoworld} reconstructs ego views from exo depth, 3D hand poses, and text via point-cloud reprojection and diffusion inpainting, while EgoExo-Gen~\cite{xu2025egoexo} uses action text and the first ego frame to guide egocentric synthesis. Intention-driven generation studies the inverse setting, exploring ego-to-exo generations~\cite{luo2024intention}. 
Exo2Ego-V~\cite{liu2024exocentric} uses multi-view exocentric inputs and camera poses within a PixelNeRF-based diffusion framework for exocentric-to-egocentric video generation. We follow the Exo2Ego-V setting, relying only on multiple exocentric views and ego/exo camera poses without ground-truth ego signals. Crucially, unlike prior methods trained from scratch or limited data, we are the first to address this task by adapting a large-scale pretrained video foundation model.

\section{Methodology}
\label{sec:method}
\subsection{Preliminaries}\label{subsec:preliminaries}
\textbf{Video Latent Diffusion Models.} 
A video Latent Diffusion Model (LDM) operates by transforming an input video into a compact latent representation through a pretrained Variational Auto-Encoder (VAE). The diffusion process is performed entirely in this latent space, where the model learns to iteratively denoise a noisy latent variable to recover a clean and coherent representation. The resulting latent is then decoded by the VAE back into the video space, yielding the final output sequence. 

The VAE used applies spatiotemporal compression with factors \((c_f, c_h, c_w) \in \mathbb{N}^3\) along the temporal and spatial dimensions, respectively, effectively reducing the number of frames by a factor of \(c_f\) and the spatial resolution by \((c_h, c_w)\). This latent-space formulation enables efficient training and inference while preserving high-fidelity video generation.

\noindent\textbf{WAN2.2 Baseline.} 
WAN2.2 is a powerful foundational video LDM designed for high-quality video generation conditioned on text and/or images. It employs a 3D Variational Auto-Encoder (VAE) with a temporal compression factor \(c_f \neq 1\), meaning that motion across every \(c_f\) consecutive frame is encoded into a single latent frame. Throughout this paper we show the temporal dimension in RGB and latent space by $F$ and $f$ respectively, where $f=\frac{F}{c_f}$. We follow the same notation for spatial dimensions where $H$ and $W$ defines the height and width in RGB space and $h=\frac{H}{c_h}$ and $w=\frac{W}{c_w}$ for latent spatial dimension.

The diffusion model component of WAN2.2 performs denoising on the video latents using a sequence of Diffusion-Transformer (DiT) blocks. Each block applies a self-attention operation over the latent video tokens, followed by a cross-attention operation between the latent video tokens and the text tokens. Image conditioning is achieved by zero-padding the input image along the temporal dimension to match the number of frames in RGB space, encoding it with the same VAE, and concatenating its latent representation with the noisy video latent. 
{Though being exposed to many training data, WAN2.2 is still not equipped with the ability to condition on a video, especially a cross-view one.}
%

\subsection{Framework Overview}
\textbf{Task Definition.} 
Given four exocentric videos 
\(\mathcal{V}^\text{exo} = \{V^\text{exo}_1, V^\text{exo}_2, V^\text{exo}_3, V^\text{exo}_4\}\) 
of shape \(4 \times F \times C \times H \times W\), captured by fixed cameras arranged 360° around daily-life skilled human activities, our goal is to generate the corresponding egocentric video 
\(V^\text{ego} \in \mathbb{R}^{F \times C \times H \times W}\).   
To model both the environment and camera motions, the model is to be conditioned on both video content and camera poses. Let the fixed exocentric camera poses be 
\(\mathcal{P}^\text{exo} = \{P^\text{exo}_1, P^\text{exo}_2, P^\text{exo}_3, P^\text{exo}_4\}\), 
where each \(P^\text{exo}_i \in \mathbb{R}^{4 \times 4}\), and let the ego camera poses be 
\(\mathcal{P}^\text{ego} = \{P^\text{ego}_t\}_{t=1}^{F}\), 
which vary across frames.  
We define the relative camera pose between the egocentric frame at time \(t\) and a reference exo camera (chosen here as the first camera) as $P^\text{rel}_t = \left(P^\text{ego}_t\right)^{-1} P^\text{exo}_1$. This formulation enables the model to jointly condition on multi-view exocentric information and temporal camera motion when generating the egocentric video.

This task is particularly challenging due to the combined effects of environmental complexity and camera motion, especially in scenarios such as bike repair. Additionally, not all exocentric views are equally informative, as some cameras may capture the operator from suboptimal angles. Therefore, the model must selectively integrate information from multiple views.

\noindent\textbf{Our Framework.} To address this problem, we propose \textit{Exo2EgoSyn}, a multi-video and camera-pose conditioned latent diffusion model that adapts the pretrained WAN2.2 model for cross-view video generation. This task is challenging for two main reasons: first, the pretrained model exhibits a bias to replicate the conditioned image as the first frame of the generated video, which complicates cross-view generation; second, WAN2.2 has not been trained on video conditions or camera poses, and its original text-based conditioning is not applicable to our setting. To overcome these challenges, {Exo2EgoSyn proposes an Ego-Exo View Alignment (EgoExo-Align), followed by a two-stage finetuning strategy on WAN2.2, which uses  Multi-view Exocentric Video Conditioning (MultiExoCon) and Pose-Aware Latent Injection (PoseInj)} to enable conditioning the model on multiple exo videos and relative camera poses we have, respectively.
The overall framework is shown in Fig. \ref{fig:fig2}

\subsection{EgoExo-Align Module}

The EgoExo-Align module is a transformer model with the purpose of predicting the latent representation of the first egocentric frame, given the latent of the first frame of each exocentric video $V^{exo}_{i}$ and the ego-exo Plücker-embedded relative camera poses. This component proved crucial, as it bypasses the tendency of the video generator to replicate the starting frame in the generated video. This effect was observed even when Wan2.2 was conditioned on a video (\textcolor{red}{SupMat}) and even when prompted to produce the egocentric video (demonstrated in Fig.~\ref{fig:fig1}). Finetuning the model proved ineffective, as the pretrained weights are already biased towards this behavior.
In contrast, by conditioning on the frame, predicted by our EgoExo-Align module, the video generator is able to predict a video with the correct viewpoint.


The EgoExo-Align module is defined as follows:
\[
\hat{l}^{\text{ego}}_1 = 
\phi\Big(
x : \big[ vae(\mathcal{V}^{\text{exo}}_{i,1}) \big]_{i=1}^{4}, \;
y : \big[ E(P^{{rel^i}}_{t=1}, K)\big]_{i=1}^{4}
\Big),
\]
where \(vae(\cdot)\) denotes the pretrained VAE, \(\mathcal{V}^{\text{exo}}_{i,1}\) represents the first frame of the \(i\)-th exocentric video, \(P^{{rel}^i}_{t=1}\) defines the relative pose of the first frame of the \(i\)-th exocentric view, and \(E(\cdot)\) applies the Plücker representation to the sparse relative poses, providing 6D per-pixel geometry-aware features given the egocentric camera intrinsic parameters \(K\). We employ STTransformer \cite{bruce2024genie}, as implemented by \cite{savov2025exploration}, denoted by $\phi$. It interleaves spatial-temporal blocks, each of which first performs attention across the height and width of the images (spatial), then temporally  across all frames (details in \textcolor{red}{SupMat}). In our case, the four exocentric frames form the temporal channels.

\subsection{{Relaxing Novel Conditions}}
As discussed earlier, our task requires a model capable of conditioning not only on a single video, but simultaneously on four videos. Moreover, the model must integrate camera poses as an additional conditioning signal. The mismatch between WAN2.2’s original capabilities and these requirements poses a substantial challenge. To address this, we introduce a {two-stage} finetuning approach: {where we first handle four exocentric video conditioning using our MultiExoCon, and then address camera pose guidance by applying our PoseInj.}

\subsubsection{{Multi-view Exocentric Video Conditioning}}
In the WAN2.2 model, each block first applies a self-attention operation over the video latent tokens, followed by a cross-attention operation with the text tokens. To enable multi-view video conditioning, we replace the text input with our four exocentric videos and train only the cross-attention modules of WAN2.2. Additionally, we employ LoRA to finetune the self-attention layers, allowing efficient adaptation to the multi-video input.

To achieve this, we first collect the VAE-encoded latents corresponding to each exocentric video, resulting in a signal of shape \((f, 4*c_\text{dim}, h, w)\). We then apply a patch embedding function \(\psi\), implemented as a 3D convolution with kernel size and stride of \((1,2,2)\), an input channel dimension of \(4*c_\text{dim}\), and an output channel dimension matching the hidden state size of the self-attention module (\(c_m\)). This operation transforms the latents into a tensor of shape \((f, c_m, \frac{h}{2}, \frac{w}{2})\).  
Next, we build a token sequence of length \(\frac{f*h*w}{4}\), with each token represented by a vector of dimension $c_m$, which we denote as \(\mathbf{X}_\text{toks}\):
\[
\mathbf{X}_\text{toks} = \textit{flatten}\Big(\psi\big([vae(\mathcal{V}^{\text{exo}}_i)]_{i=1}^{4}\big)\Big),
\]
where \(\text{flatten}(\cdot)\) is the flattening operator to produce the token sequence. These tokens are then fed into the cross-attention module to update the hidden states \(\mathbf{h}\):
\[
\mathbf{h} \leftarrow \mathbf{h} + \textit{CA}(\mathbf{X}_\text{toks}, \mathbf{h}).
\]

Importantly, in this stage, the output of \(\phi(\cdot)\) consistently serves as the conditioning input to the diffusion model.

\subsubsection{{Pose-Aware Latent Injection}}
Sherwin et al.~\cite{bahmani2025ac3d} demonstrate that camera-pose information constitutes a low-frequency signal that can easily be overlooked by large DiT-based diffusion models. Consequently, simply injecting pose information into an intermediate layer is insufficient for effective conditioning. Following the approach of \cite{bahmani2024vd3d}, we first convert the sparse relative camera poses into a dense per-pixel representation, given the intrinsic matrix ($K$) of the ego camera. This is achieved using Plücker representation, which encodes each pixel with a 6-dimensional feature vector uniquely defining a 3D ray. The resulting tensor has a shape ($F \times 6 \times H \times W$).

The Plücker embeddings are computed for each frame, yielding \(F\) representations, whereas WAN2.2 operates on a temporally compressed latent space, where every four consecutive frames are encoded into a single latent ($c_f=4$). As a result, the raw Plücker coordinates cannot be directly incorporated into the model. While one could naively downsample or reshape the pose sequence to match the latent resolution, this approach would inefficiently merge motion information across the four-frame window.  

{Instead,} following \cite{bahmani2024vd3d}, we apply a sequence of causal 1D convolutions to transform the \(F \times 6\) Plücker coordinate sequence for each pixel into a \((f \times 96)\) representation for each \(8 \times 8\) spatial grid. This produces a tensor of shape \(f \times 96 \times H/8 \times W/8\), which is then concatenated to the hidden state prior to entering the first DiT block of WAN2.2. We observe that the pretrained weights obtained from the multi-video cross-attention finetuning stage facilitate effective conditioning on camera poses in this step. Since the camera-pose features are fed into the DiT blocks, we also employ LoRA finetuning for the self-attention modules to adapt them to the pose-conditioned input. {We highlight that although pose embeddings have been explored in prior work, this is the first time they are introduced into the Exo2Ego generation task, through a natural and effective modification that embeds relative exo-to-ego camera poses.}

\subsection{{Training and Loss Functions}}

We first train the ExoEgoAlign module, minimizing \[
\mathcal{L}_1 = \big\| l^{\text{ego}}_1 - \hat{l}^{\text{ego}}_1 \big\|_1
,\]
and use the detached produced latents as the condition to WAN2.2. We then train the cross-attention layers with the video tokens from the four exocentric videos. We apply LoRA finetuning to the self-attention layers, allowing the model to leverage motion information directly induced by the video frames to better adapt the target egocentric frames. Specifically, we optimize the diffusion objective:
\[
\mathcal{L}_2 = \Big\| \epsilon_t - \mathcal{W}\big([z_t, \hat{l}^{\text{ego}}_1], \mathcal{V}^\text{exo}, t \big) \Big\|^2
,\]
where \(z\) denotes the noisy latent input, \(\mathcal{W}\) indicates the WAN2.2 diffusion model, \(t\) is the time step, and \(\epsilon\) is the ground-truth noise applied in the forward diffusion process.


In the second stage, we initialize the model with the weights obtained from the previous stage and concatenate the Plücker embeddings to the hidden state before feeding it into the sequence of DiT blocks. During this stage, we LoRA finetune the self-attention modules and optimize the following loss:
\[
\mathcal{L}' = \Big\| \epsilon_t - \mathcal{W}\big([z_t, \hat{l}^{\text{ego}}_1, E(P^{rel}, K)], \mathcal{V}^\text{exo}, t \big) \Big\|^2
.\]

Each stage is trained for 100k steps on 2×NVIDIA H200 GPUs, with a training time of around 10 hours. More details on model size, training configurations, and hyperparameters are provided in the \textcolor{red}{SupMat}.

\section{Experiments}
\label{sec:experiment}









\begin{table}[t]
  \caption{\textbf{Comparative results.} Exo2EgoSyn Results vs. the Baseline VAWAN}
  \label{tab:baseline_comparison}
  \begin{center}
    \begin{small}
      \begin{sc}
        \begin{tabular}{lccc}
          \toprule
          Category & PSNR$\uparrow$ & SSIM$\uparrow$ & LPIPS$\downarrow$ \\
          \midrule

          \textbf{Bike} \\
          \quad \textit{VAWAN} &
          14.6352 & 0.3516 & 0.5553 \\
          \quad \textit{Ours} &
          \textbf{15.1319} & \textbf{0.3895} & \textbf{0.5052} \\[3pt]

          \textbf{Basketball} \\
          \quad \textit{VAWAN} &
          15.9996 & 0.4768 & 0.4989 \\
          \quad \textit{Ours} &
          \textbf{16.6445} & \textbf{0.5052} & \textbf{0.4158} \\[3pt]

          \textbf{CPR} \\
          \quad \textit{VAWAN} &
          15.2329 & 0.4935 & 0.5249 \\
          \quad \textit{Ours} &
          \textbf{16.7229} & \textbf{0.5294} & \textbf{0.4096} \\[3pt]

          \textbf{Covid} \\
          \quad \textit{VAWAN} &
          14.0127 & 0.3889 & 0.5974 \\
          \quad \textit{Ours} &
          \textbf{14.4812} & \textbf{0.4312} & \textbf{0.4868} \\

          \bottomrule
        \end{tabular}
      \end{sc}
    \end{small}
  \end{center}
  \vskip -0.1in
\end{table}

\begin{figure*}[t]
  \centering
  \includegraphics[width=\linewidth]{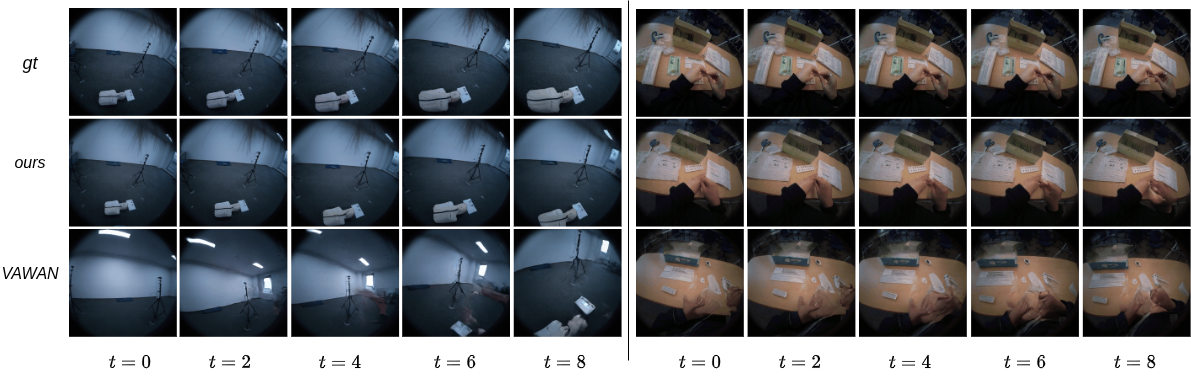}
  \caption{\textbf{Qualitative Results.} Left: a scene from the CPR subcategory; Right: a scene from the Covid Test subcategory.}
  \label{fig:figx}
  \vskip -0.2in
\end{figure*}

\subsection{Experimental Setup}
\noindent\textbf{Dataset.}
We evaluate our approach in four categories of the Ego-Exo4D~\cite{grauman2024ego} benchmark: Bike, Basketball, Covid Test, and CPR. For each egocentric video, the dataset provides four exocentric videos along with the corresponding camera poses for the fixed exocentric cameras and for every frame of the egocentric video. Additionally, the intrinsic parameters of the egocentric camera are provided, which we utilize in the Plücker embedding computation.  
Our experiments process 363 videos from the Bike category, 910 videos from Basketball, 246 videos from Covid Test, and 98 videos from CPR. Both egocentric and exocentric videos are spatially resized to a resolution of \(H = W = 256\). Since the WAN2.2 architecture requires the number of frames to satisfy \(4n + 1\), we sample 9 frames per video{, one more than the 8-frame architecture used in Exo2Ego-V~\cite{liu2024exocentric}}. Using a skip-frame step of 4 during preprocessing ensures sufficient motion is captured in the sampled frames. For training and testing splits, we follow the official partitions released by \cite{grauman2024ego}. {Empirically, as partly shown in Table~\ref{tab:recammaster_comparison_ablation}, our framework is sensitive to conditioning on multiple exocentric views, and replacing exo video tokens with text tokens in cross-attention consistently improves qualitative results. To enable conditioning on both the first egocentric frame and multiple exocentric views, we adopt the Image-to-Video (I2V) model from WAN2.2 rather than the Text-to-Video (T2V-5B) model, which lacks first-frame conditioning.}


\noindent\textbf{Baseline Construction.}\label{VAWAN}
{The Perception-Distortion tradeoff~\cite{blau2018perception} shows that optimizing for distortion often comes at the expense of perceptual quality. Current state-of-the-art methods~\cite{liu2024exocentric}, built on smaller pretrained models, focus on distortion, while foundation models like ours prioritize perceptual realism. Comparing the two via distortion metrics (e.g., PSNR) is therefore inherently unfair. To enable a fair evaluation, we propose a WAN-based baseline, which, like our method, falls in the perception-focused category.} As discussed in Section~\ref{subsec:preliminaries}, WAN2.2  natively supports video-shaped input, since it zero-pads the conditioned image to match the temporal dimension. Leveraging this capability, we introduce a baseline, \textit{Video-Accepting WAN (VAWAN)}, for our experiments. In VAWAN, the model is conditioned on four exocentric videos by generalizing the standard image-conditioning mechanism: rather than conditioning on a single frame, we concatenate the VAE-encoded latents of all four exocentric videos along the channel dimension and integrate them with the noisy input latent, preserving the original architectural design. To incorporate camera motion, we evaluate both raw camera poses and dense Plücker embeddings under various configurations, as detailed in our ablation study.

\noindent\textbf{Evaluation Metrics.}
To quantitatively evaluate our method, we assess performance using PSNR, SSIM and LPIPS (AlexNet~\cite{krizhevsky2012imagenet}), following the evaluation protocol outlined in \cite{liu2024exocentric} to ensure a consistent and fair comparison with the baseline. However, because these metrics do not fully capture the perceptual quality of generative output, we also conduct a user study. A total of 100 video samples—25 from each category—are evaluated by human participants, who compare the outputs of our model against those of the baseline. More details on user study are provided in the \textcolor{red}{SupMat}.

\subsection{Comparison with Baseline}
As summarized in Table~\ref{tab:baseline_comparison}, our model consistently outperforms VAWAN in all four categories, with particularly notable gains in SSIM and LPIPS. On average, we observe an approximate 10\% improvement in SSIM and a 16\% reduction in LPIPS, indicating substantial enhancements in perceptual quality. See qualitative results in Fig. \ref{fig:figx}. Please refer to the \textcolor{red}{SupMat} for more qualitative results.

\begin{figure}[t]
  \centering

  \includegraphics[width=0.5\linewidth]{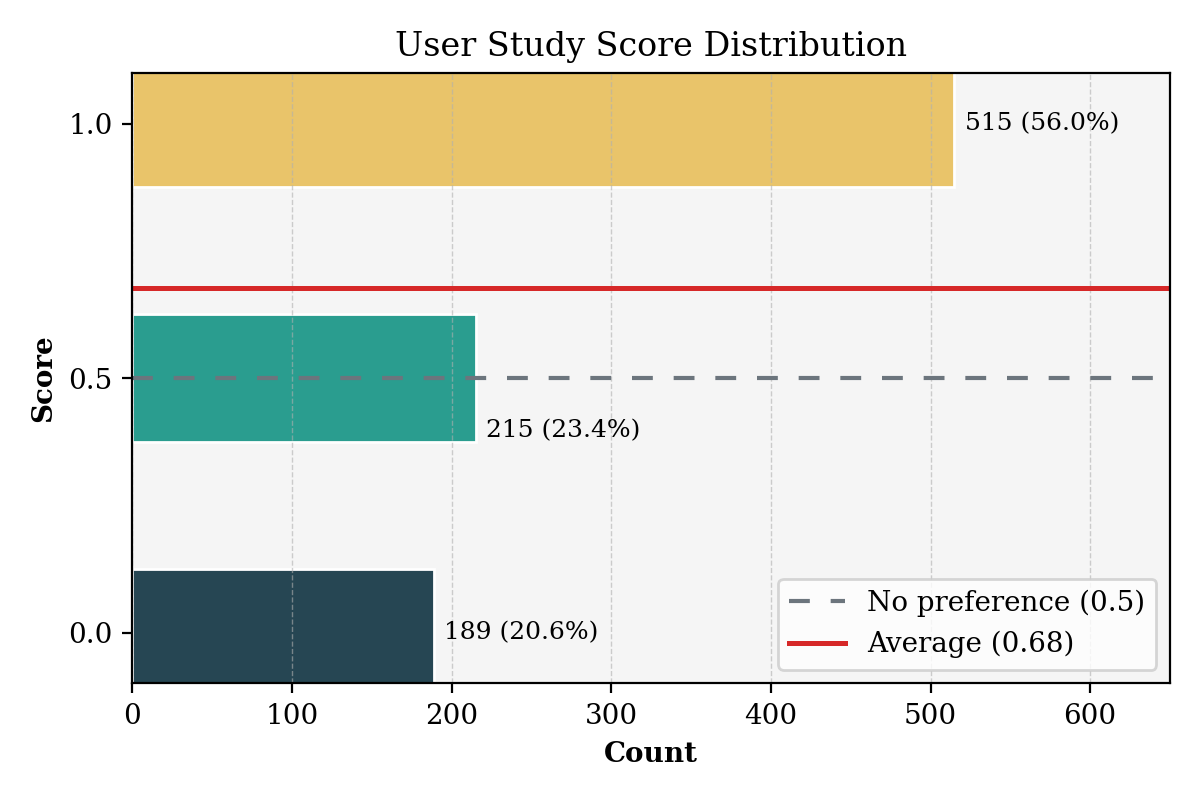}
  \caption{\textbf{User Study Score Distribution.} A clear preference to our model (score 1) is noticed, compared to the lower preference votes for baseline (score 0) or no preference (score 0.5).}
  \label{fig:user_study}
  \vskip -0.2in
\end{figure}

To further demonstrate the effectiveness of our pipeline, we conducted a complementary user study with 10 participants (81 out of 1000 questions were skipped by them), following the design of \cite{savov2025statespacediffuser}. Scoring is based on the ground truth similarity of our model's outputs against those of the baseline VAWAN, in terms of hand-object interaction, camera motion and visual fidelity. The scoring options are preference for ours (1.0), for the baseline (0.0), or no preference (0.5). Fig.~\ref{fig:user_study} shows the preference distribution, indicating a clear preference for our model, also confirmed by a final average score of \textbf{0.68}.  
\subsection{Ablation Studies}

We perform an ablation study to evaluate the contributions of different components in our pipeline. Specifically, we investigate the presence and architectural variations of the EgoExo-Align module. We also assess the role of the proposed PoseInj module during the second stage of fine-tuning. {For a fair comparison with other strong conditioning mechanisms, we additionally evaluate our method against ReCamMaster \cite{bai2025recammaster}}. Finally, we examine the impact of the number of exocentric views and various camera-pose conditionings. {We also perform an ablation on the number of diffusion inference steps and compare with Exo2Ego-V~\cite{liu2024exocentric} in the \textcolor{red}{SupMat}.}

\paragraph{EgoExo-Align Module.}
As shown in Table~\ref{tab:multiexocon_bike}, we ablate the presence of the EgoExo-Align module and also establish an upper bound for its performance by directly using the first frame of the target ego video. We observe that removing this module leads to a notable degradation in performance. All experiments are conducted with the proposed MultiExoCon module, but excluding the PoseInj module, to isolate and evaluate the effect of EgoExo-Align.

As shown in Table~\ref{tab:performance_bike}, we further ablate the effectiveness of different architectures used to implement the EgoExo-Align module, with the STTransformer achieving the most promising results. For these experiments, we use the proposed baseline, VAWAN, under various camera-pose conditioning scenarios. When active, camera poses are embedded into the model depending on the architecture; specifically, the STTransformer leverages the Plücker coordinates of the relative camera poses from the first frames of each exocentric video.


\begin{table}[t]
  \caption{\textbf{EgoExo-Align Existence.} \textit{GT Frame} conditions on the first ground-truth ego frame, providing an upper-bound performance for our proposed module.}
  \label{tab:multiexocon_bike}
  \begin{center}
    \begin{small}
      \begin{sc}
        \renewcommand{\arraystretch}{1.0}
        \resizebox{0.8\columnwidth}{!}{%
        \begin{tabular}{l l c c c}
          \toprule
           & \textbf{Scenario} & \multicolumn{3}{c}{\textbf{Bike}} \\
          \cmidrule(lr){3-5}
           &  & PSNR↑ & SSIM↑ & LPIPS↓ \\
          \midrule

          \multirow{3}{*}{\textit{MultiExoCon}}
          & +EgoExo-Align
          & 15.0216
          & 0.3798
          & 0.5101 \\

          & -EgoExo-Align
          & 14.7361
          & 0.3450
          & 0.5359 \\

          & +GT frame
          & 15.5327
          & 0.4321
          & 0.3951 \\

          \bottomrule
        \end{tabular}}
      \end{sc}
    \end{small}
  \end{center}
  \vskip -0.3in
\end{table}







\begin{table}[t!]
  \caption{\textbf{EgoExo-Align Architecture.} The \textit{Cross-Attn} model uses convolutional layers to encode the input frames, and the applies a cross-attention module between these merged encodings and the camera-pose embeddings. We use the pretrained \textit{Pixel-Nerf} models for different categories from \cite{liu2024exocentric}. The superscripts $S$ and $L$ indicate the size of the spatiotemporal transformer: the small version ($S$) has 4 heads, 4 blocks, and a dimension of 128, while the large version ($L$) has 8 heads, 8 blocks, and a dimension of 512.}
  \label{tab:performance_bike}
  \begin{center}
    \begin{small}
      \begin{sc}
        \renewcommand{\arraystretch}{1.0}
        \resizebox{0.8\columnwidth}{!}{%
        \begin{tabular}{l l c c c c}
          \toprule
          \textbf{Base} & \textbf{Arc} & \textbf{Pose} &
          \multicolumn{3}{c}{\textbf{Bike}} \\
          \cmidrule(lr){4-6}
           & & & PSNR↑ & SSIM↑ & LPIPS↓ \\
          \midrule

          \multirow{5}{*}{\textit{VAWAN}}
          & Cross-Attn & N 
          & 14.1216 
          & 0.3342 
          & 0.5988 \\

          & Cross-Attn & Y 
          & 14.2902 
          & 0.3391 
          & 0.5773 \\

          & Pixel-Nerf & Y 
          & 14.5223 
          & 0.3301 
          & 0.5944 \\

          & ST-Transformer$^{S}$ & Y 
          & 14.5387 
          & 0.3529 
          & 0.5470 \\

          & ST-Transformer$^{L}$ & Y 
          & 14.7227 
          & 0.3630 
          & 0.5320 \\

          \bottomrule
        \end{tabular}}
      \end{sc}
    \end{small}
  \end{center}
  \vskip -0.3in
\end{table}







\begin{table}[t!]
  \caption{\textbf{PoseInj Presence.} \textit{gt-frame} indicates that the EgoExo-Align module is not used, and the model is directly conditioned on the first frame of the target ego video.}
  \label{tab:poseinj_comparison}
  \begin{center}
    \begin{small}
      \begin{sc}
        \renewcommand{\arraystretch}{1.5}
        \resizebox{0.7\textwidth}{!}{%
        \begin{tabular}{l|ccc|ccc}
          \toprule
          \textbf{Category} &
          \multicolumn{3}{c|}{\textbf{ours w/o PoseInj}} &
          \multicolumn{3}{c}{\textbf{ours}} \\
          \cmidrule(lr){2-4} \cmidrule(lr){5-7}
           & PSNR↑ & SSIM↑ & LPIPS↓ & PSNR↑ & SSIM↑ & LPIPS↓ \\
          \midrule

          Bike (gt-frame) &
          15.5327 &
          0.4321 &
          0.3951 &
          15.5862 &
          0.4326 &
          0.3881 \\

          Bike &
          15.0216 &
          0.3798 &
          0.5101 &
          15.1319 &
          0.3895 &
          0.5052 \\

          Basketball &
          16.3158 &
          0.4901 &
          0.4299 &
          16.6445 &
          0.5052 &
          0.4158 \\

          CPR &
          15.9145 &
          0.4617 &
          0.4308 &
          16.7229 &
          0.5294 &
          0.4096 \\

          Covid &
          14.0823 &
          0.4091 &
          0.5103 &
          14.4812 &
          0.4312 &
          0.4868 \\

          \bottomrule
        \end{tabular}}
      \end{sc}
    \end{small}
  \end{center}
  \vskip -0.4in
\end{table}









\paragraph{PoseInj Module.}
As illustrated in Table~\ref{tab:poseinj_comparison}, we conduct an extensive ablation on the presence of the PoseInj component during the second fine-tuning step across all categories of the Ego-Exo4D dataset. The results demonstrate the robustness of the proposed module across different experimental setups.
\paragraph{{Benchmarking Against ReCamMaster.}}
{ReCamMaster introduces a camera-controlled framework to render a target video from a source video. Specifically, it leverages a frame-concatenation technique between the conditioning and actual video signals. As shown in Table~\ref{tab:recammaster_comparison_ablation}, we compare our results with ReCamMaster across two different model sizes; 5B and 14B. While the larger model achieves a higher PSNR on the Bike subcategory, it falls significantly behind our method in LPIPS and SSIM.}

\begin{table}[h]
  \caption{\textbf{Comparison with ReCamMaster~\cite{bai2025recammaster}.} To remain consistent with the ReCamMaster (RCM) design, we ran it using the 5B and 14B Text2Video (T2V) backbones, which are adapted from WAN2.2.}
  \label{tab:recammaster_comparison_ablation}
  \begin{center}
    \begin{small}
      \begin{sc}
        \resizebox{0.55\columnwidth}{!}{%
        \begin{tabular}{lccc}
          \toprule
          Framework & \multicolumn{3}{c}{Bike} \\
          \cmidrule(lr){2-4}
           & PSNR↑ & SSIM↑ & LPIPS↓ \\
          \midrule
          RCM-T2V-5B   & 14.9651 & 0.3537 & 0.5737 \\
          RCM-T2V-14B  & \textbf{15.2017} & 0.3598 & 0.5680 \\
          Ours                  & 15.1319 & \textbf{0.3895} & \textbf{0.5052} \\
          \bottomrule
        \end{tabular}}
      \end{sc}
    \end{small}
  \end{center}
  \vskip -0.5in
\end{table}



\paragraph{Exo Views.}
We ablate the number of exocentric videos used for generation. As expected and shown in Table~\ref{tab:exo_number_ablation}, due to the limited training data compared to what the WAN2.2 model was trained on, training the full model can lead to overfitting. Furthermore, increasing the number of exocentric videos generally improves performance, as some videos may be captured from suboptimal angles. Here, we incorporate the pose-conditioning approach proposed in \cite{bahmani2025ac3d} into our baseline model.

\begin{table}[t!]
  \caption{\textbf{Number of Exo Views Used.} When using 1 exo video, we randomly choose between 4 available views.}
  \label{tab:exo_number_ablation}
  \begin{center}
    \begin{small}
      \begin{sc}
        \renewcommand{\arraystretch}{1.0}
        \resizebox{0.65\columnwidth}{!}{%
        \begin{tabular}{l c l c c c}
          \toprule
          \textbf{Scenario} & \textbf{V} & \textbf{Training} &
          \multicolumn{3}{c}{\textbf{Bike}} \\
          \cmidrule(lr){4-6}
           & & & PSNR↑ & SSIM↑ & LPIPS↓ \\
          \midrule

          \multirow{3}{*}{\textit{VAWAN-AC3D}}
          & 1 & LoRA
          & 14.5278
          & 0.3375
          & 0.5821 \\

          & 4 & LoRA
          & 14.8511
          & 0.3709
          & 0.5289 \\

          & 4 & Whole
          & 14.5101
          & 0.3305
          & 0.5971 \\

          \bottomrule
        \end{tabular}}
      \end{sc}
    \end{small}
  \end{center}
  \vskip -0.2in
\end{table}













\begin{table}[t!]
  \caption{\textbf{Camera Conditioning.} \textit{Time\&Pose} is adapted from \cite{liu2024exocentric}, where time steps and camera-pose information are first fused before being input to the diffusion model. The numbers in parentheses ($\cdot$) indicate the bottleneck dimension of the additional DiT blocks proposed in \cite{bahmani2025ac3d}.}
  \label{tab:pose_ablation}
  \begin{center}
    \begin{small}
      \begin{sc}
        \renewcommand{\arraystretch}{1.0}
        \resizebox{0.65\columnwidth}{!}{%
        \begin{tabular}{l c c c c}
          \toprule
          \multicolumn{5}{c}{\textit{VAWAN\; gt-frame}} \\
          \midrule
          \textbf{Pose cond} & \textbf{V} & PSNR↑ & SSIM↑ & LPIPS↓ \\
          \midrule

          Time\&Pose & 1 
          & 14.2125 
          & 0.3252 
          & 0.5944 \\

          AC3D (512) & 1 
          & 15.2965 
          & 0.4350 
          & 0.4027 \\

          AC3D (512) & 4 
          & 15.2524 
          & 0.4346 
          & 0.4057 \\

          AC3D (512) Shared & 4 
          & 15.1409 
          & 0.4298 
          & 0.4169 \\

          AC3D (2048) & 4 
          & 15.3268 
          & 0.4368 
          & 0.4026 \\

          \bottomrule
        \end{tabular}}
      \end{sc}
    \end{small}
  \end{center}
  \vskip -0.3in
\end{table}

\paragraph{Camera Conditioning.}
Here, as shown in Table \ref{tab:pose_ablation}, we investigate different strategies for camera-pose conditioning, following the approaches of \cite{liu2024exocentric} and \cite{bahmani2025ac3d}. The former uses a UNet-based architecture and fuses information between time steps and camera poses, while the latter introduces additional DiT blocks, applying an attention mechanism between compressed latent representations and camera poses in a lower-dimensional space. We ablate different bottleneck dimensions for this compression and also evaluate a variant in which the newly added DiT blocks are shared across all main DiT blocks of WAN2.2.

\section{Conclusion}
\label{sec:conclusion}

We propose a novel framework to adapt strong foundational video generation models to the Exo2Ego cross-view translation task. The task presents two key challenges: (1) breaking the biases learned by the foundation model, and (2) aligning the conditioning input with the model’s expected format. To address the first, we introduce \textbf{EgoExo-Align}, which predicts the latent of the target ego frame from the first frames of multiple exocentric views. For the second, we employ a two-stage adaptation: the first stage conditions WAN2.2 on all exocentric videos to provide informative cues about future ego frames by use of the \textbf{MultiExoCon}, while the second stage enhances camera-tracking by incorporating Plücker embeddings of relative camera poses into the hidden state before the DiT blocks, using \textbf{PoseInj} module. We also introduce a baseline, {Video-Accepting WAN (VAWAN)}, which extends WAN2.2 to accept multi-video conditions and explore different camera-pose conditioning scenarios. Our quantitative and qualitative evaluations on the challenging subcategories of the Ego-Exo4D dataset demonstrate the robustness of our approach over the VAWAN baseline model.

\section{Acknowledgements}
This research was partially funded by the Ministry of Education and Science of Bulgaria (support for INSAIT, part of
the Bulgarian National Roadmap for Research Infrastructure), and the dAIedge project (HORIZON-CL4-2022-HUMAN-02-02, Grant Agreement Number: 101120726).

\bibliographystyle{splncs04}
\bibliography{main}

\onecolumn

\title{Exo2EgoSyn Supplementary Materials} 
\titlerunning{Exo2EgoSyn}

\author{Mohammad Mahdi\inst{1} \quad Yuqian Fu\inst{1} \quad Nedko Savov\inst{1} \quad Jiancheng Pan\inst{1} \quad \quad
Danda Pani Paudel\inst{1}  \quad Luc Van Gool\inst{1}}

\authorrunning{Mahdi et al.}

\institute{$^1$INSAIT, Sofia University “St. Kliment Ohridski” \\
\texttt{firstname.lastname@insait.ai}}

\maketitle

\section{Biases in Video Foundation Models}
As outlined in the Introduction, obstacle \ding{172}: \textbf{Strong view bias}, and in Section 3.2 (\textbf{Our Framework}), foundation video generative models tend to reproduce the conditioned image as the first frame of the generated video. This behavior significantly complicates our goal of cross-view generation. In this section, we present examples demonstrating how the WAN2.2 \cite{wan2025wan} model consistently prioritizes the conditioned image over the conditioned text—even in cases where we explicitly prompt the model to avoid replicating the first frame. We first examine this effect using general, non-task-specific text prompts, and then with ego-centric prompts tailored to our setting. Across all cases, the model exhibits a strong bias toward preserving the first frame, indicated by red in Fig. \ref{fig:figs1}.
\definecolor{myblue}{RGB}{0, 100, 200}
\definecolor{mygreen}{RGB}{34, 139, 34}
\begin{figure*}[t]
  \includegraphics[width=\linewidth]{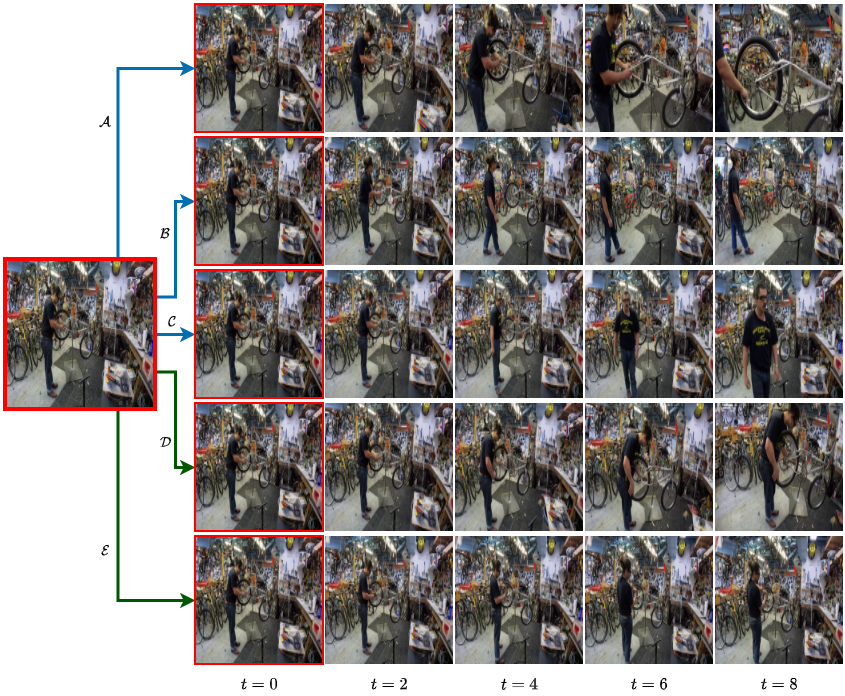}
  \caption{\textbf{WAN2.2's Bias of Replicating the Conditioned Image.} On the left is the conditioning image. On the right, WAN2.2 generates five distinct videos based on different text prompts. The first three (blue arrows) use general prompts, while the last two (green arrows) use prompts tailored for our task. The text prompts used are as follows: \textcolor{myblue}{\underline{$\mathcal{A}$}}: \textit{"Do start the video from a black frame."}, \textcolor{myblue}{\underline{$\mathcal{B}$}}: \textit{"Begin the video at a moment when the man is already in motion farther down the bike, as if some time has passed since the moment shown in the image."}, \textcolor{myblue}{\underline{$\mathcal{C}$}}: \textit{"The video must not begin with a scene resembling the supplied image. The opening frame should show a noticeably changed situation — the man in a different location or pose."}, \textcolor{mygreen}{\underline{$\mathcal{D}$}}: \textit{"Produce a fully ego-centric video starting from the provided image. The camera must be strictly from the man’s eyes, showing the environment as he sees it — hands, body, and movement should appear naturally from a first-person perspective."}, and \textcolor{mygreen}{\underline{$\mathcal{E}$}}: \textit{"Generate the ego-centric video from the image.".}}
  \label{fig:figs1}
\end{figure*}

\section{Impact of the EgoExo-Align Module}
\begin{figure*}[p]
  \includegraphics[width=\linewidth]{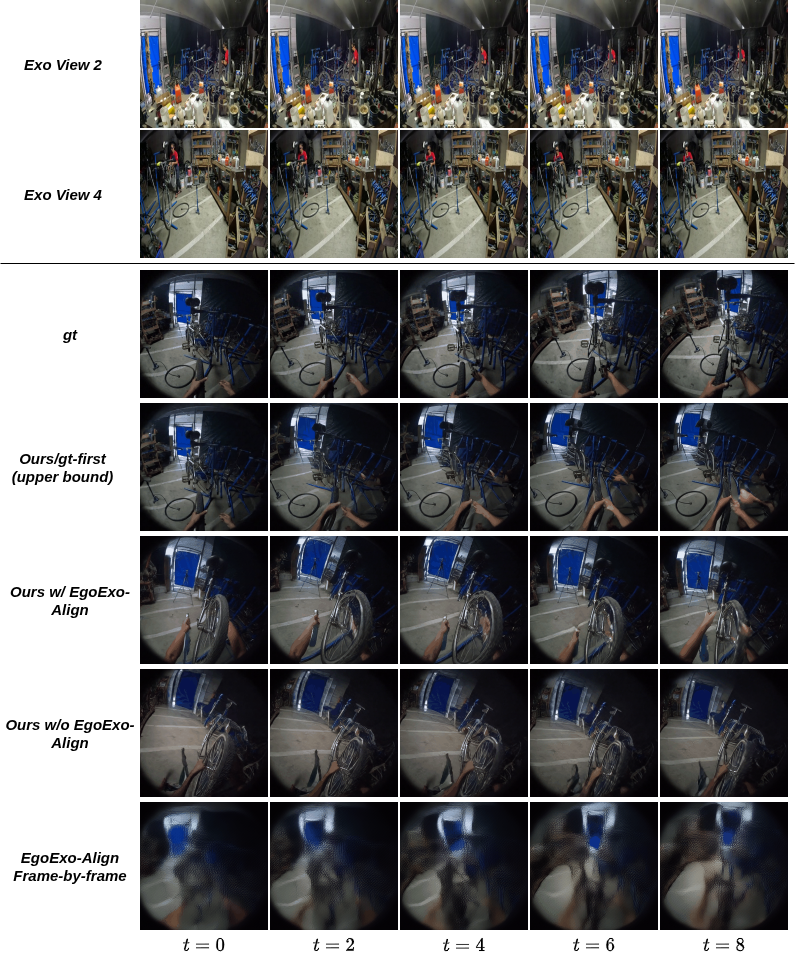}

  \caption{\textbf{EgoExo-Align Module.} Top: Model inputs. The second and fourth exo videos are shown for illustration, appearing in the first and second rows, respectively. Bottom: The second row presents an upper-bound performance by conditioning Exo2EgoSyn on a ground-truth ego-view first frame. A comparison between the third and fourth rows—with and without the EgoExo-Align module, respectively—confirms the module's necessity. Conversely, the last row, which results from applying the module in a standalone, frame-by-frame manner, demonstrates its insufficiency. This contrast highlights that effective synthesis requires both the EgoExo-Align module and the subsequent fine-tuning of the WAN2.2 model (third row).}
  \label{fig:figs2}
\end{figure*}

Our ablation studies (Section 4.3, Table 2) confirm the necessity of the EgoExo-Align module, as its removal causes a significant performance drop. An upper performance bound is established when the model is provided with the ground-truth first frame from the ego-view video. To further isolate this module's contribution, we applied EgoExo-Align in a frame-by-frame manner (Fig. \ref{fig:figs2}, last row). The resulting poor performance demonstrates that, while necessary, the module is insufficient on its own and produces blurry outputs. It is the subsequent fine-tuning with WAN2.2 that is critical for transforming this initial alignment into a high-quality video, confirming their synergistic relationship.

\section{Training Setup and Hyperparameters}
In this section, 
we first describe the architecture of the EgoExo-Align module, the WAN2.2 VAE, and the WAN2.2 denoising transformer. Subsequently, we provide the specific training parameters for each stage of the pipeline.
\subsection{Models}

\subsubsection{{EgoExo-Align Module}}
Our EgoExo-Align module is implemented as a non-causal STTransformer \cite{bruce2024genie}, with the hyperparameters in Table~\ref{tab:sttransformer}. Attention is applied in a sequential order: first spatial, then temporal. The module produces 22 output channels, of which 16 correspond to latent VAE features and 6 to Plücker embeddings. Within the architecture, we employ the Position Encoding Generator (PEG) \cite{chu2021conditional} in its non-causal configuration.

\begin{table}[t]
\caption{\textbf{Hyperparameters of the EgoExo-Align Module.}}
\centering
\small
\begin{tabular}{l l}
\hline
\textbf{Parameter} & \textbf{Value} \\
\hline
\multicolumn{2}{l}{\textbf{STTransformer}} \\
Out channels & 22 \\
Dim & 512 \\
Heads & 8 \\
Num blocks & 8 \\
Dim head & 64 \\
FF multiplier & 4.0 \\
Attn dropout & 0.0 \\
FF dropout & 0.0 \\
\hline
\end{tabular}
\label{tab:sttransformer}
\end{table}

\subsubsection{WAN2.2 VAE}
WAN2.2 provides five main sets of pre-trained weights:
\begin{itemize}
 \item \textit{T2V-A14B}: Text2Video Mixture-of-Experts(MoE) model which supports 480/720P video generation.
 \item \textit{I2V-A14B}: Image2Video MoE model which supports 480/720P video generation.
 \item \textit{TI2V-5B}: T2V+I2V, using high-compression VAE which supports 720P video generation.
 \item  \textit{S2V-14B}: Speech2Video model which supports 480/720P video generation.
 \item \textit{Animate-14B}: character animation and replacement. 
\end{itemize}

The first two models use the VAE from WAN2.1 \cite{wan2025wan}, while the third introduces a new VAE with higher spatio-temporal compression. As defined in Section 3.1, the compression factors $(c_f, c_h, c_w)$ are $(4, 8, 8)$ for the first two setups and $(4, 16, 16)$ for the third. In this work, we use the I2V-A14B model, which employs the VAE with parameters $c_f=4$, $c_h=8$, $c_w=8$. This VAE is frozen during all experiments and contains 127M parameters.

\subsubsection{WAN2.2 Transformer}
As previously stated, our work is based on the I2V-A14B model. This model employs a mixture of two experts: a high-noise transformer for denoising at timesteps $t > \theta$ (low SNR) and a low-noise transformer for timesteps $t \leq \theta$ (high SNR). The value of $\theta=900$ in the actual implementation. These two experts are identical in architecture, each having 14B parameters. We use only the low-noise transformer for all experiments in this work. The hyperparameters used for the transformer are detailed in Table~\ref{tab:tabs2}.

\begin{table}[t]
\caption{\textbf{Configuration of the Denoising Transformer.}}
\centering
\small
\begin{tabular}{l l}
\hline
\textbf{Parameter} & \textbf{Value} \\
\hline
\multicolumn{2}{l}{\textbf{VAE}} \\
VAE Kind & Wan2.1\_VAE \\
VAE stride & (4, 8, 8) \\
\hline
\multicolumn{2}{l}{\textbf{Transformer}} \\
Patch size & (1, 2, 2) \\
Dim & 5120 \\
FFN dim & 13824 \\
Freq dim & 256 \\
Num heads & 40 \\
Num DiT blocks & 40 \\
QK norm & True \\
Cross-attn norm & True \\
Epsilon & $1\times10^{-6}$ \\
Total timesteps & 1000 \\
\hline
\multicolumn{2}{l}{\textbf{Inference}} \\
Sample shift & 5.0 \\
Boundary ($\theta$) & 900 \\
Sample guide scale & 3.5 \\
Inference steps & 40 \\
\hline
\end{tabular}
\label{tab:tabs2}
\end{table}

\subsection{Training}

\subsubsection{{EgoExo-Align Module}}
We train the STTransformer model, which has 50M parameters, using the training configurations listed in Table~\ref{tab:egoexo-align-training}. {The idea of generating the first egocentric frame could, in principle, be extended to produce the entire video. However, this approach has two major drawbacks. First, it would require as many forward passes as there are frames, making inference significantly less efficient. Second, predicting frames independently would result in poor motion consistency, necessitating an additional stage to fine-tune a motion module. Therefore, the primary purpose of this module is to mitigate the strong view bias of WAN2.2, rather than to generate the full video. For this reason, we only predict the first frame of the target egocentric video.}

\begin{table}[t]
\caption{\textbf{Training parameters of the EgoExo-Align Module.}}
\centering
\small
\begin{tabular}{l l}
\hline
\textbf{Parameter} & \textbf{Value} \\
\hline
\multicolumn{2}{l}{\textbf{Core Training}} \\
Train batch size & 128 \\
Mixed precision & bf16 \\
Max grad norm & 1.0 \\
Loss function & L1 \\
Views (exo) & 4 \\
\hline
\multicolumn{2}{l}{\textbf{Optimization}} \\
Learning rate & $1\times10^{-4}$ \\
Scheduler & cosine \\
Warmup steps & 200 \\
Min learning rate & $5\times10^{-5}$ \\
Optimizer & Adam \\
Weight decay & $1\times10^{-4}$ \\
Adam $\epsilon$ & $1\times10^{-8}$ \\
Adam betas & (0.9,\; 0.95) \\
\hline
\end{tabular}
\label{tab:egoexo-align-training}
\end{table}

\subsubsection{MultiExoCon Module}
The objectives for this fine-tuning stage are twofold: first, to enable a holistic understanding of the environment from the multi-view exocentric inputs, and second, to learn camera motion implicitly from the visual dynamics of the frames, without relying on explicit camera poses:
\[
\mathcal{L}_2 = \Big\| \epsilon_t - \mathcal{W}\big([z_t, \hat{l}^{\text{ego}}_1], \mathcal{V}^\text{exo}, t \big) \Big\|^2.
\]

As outlined in the paper (Line 335), when conditioning on four exocentric videos, the detached output of the EgoExo-Align module ($\hat{l}^{\text{ego}}_1$) serves as the consistent input to the diffusion model. We detail the training setup for this stage in Table~\ref{tab:tabs4}. In this stage, we train all cross-attention parameters in each DiT \cite{peebles2023scalable} block, while using LoRA \cite{hu2022lora} to fine-tune only the self-attention modules. The LoRA components introduce 210M trainable parameters. For the inference stage, we follow the same configuration settings described in Table~\ref{tab:tabs2}.

\begin{table}[t]
\caption{\textbf{Training Setup.} Both the first (MultiExoCon module) and second (PoseInj module) fine-tuning stages use the same training setup, differing only in the set of trainable parameters.}
\centering
\small
\begin{tabular}{l l}
\hline
\textbf{Setting} & \textbf{Value} \\
\hline
LoRA rank & 128 \\
LoRA alpha & 128 \\
Precision & bf16 \\
Exo size (H$\times$W) & 256 $\times$ 256 \\
Ego size (H$\times$W) & 256 $\times$ 256 \\
FPS & 3 \\
Frames & 9 \\
Batch size & 3 \\
Training steps & 100000 \\
Grad. accumulation & 1 \\
Learning rate & $1\times10^{-4}$ \\
Min learning rate & $5\times10^{-5}$ \\
LR scheduler & cosine \\
Warmup steps & 200 \\
LR cycles & 1 \\
Optimizer & Adam (8-bit) \\
Weight decay & $1\times10^{-4}$ \\
Adam $\epsilon$ & $1\times10^{-8}$ \\
Adam $\beta_1$ & 0.9 \\
Adam $\beta_2$ & 0.95 \\
Max grad norm & 1.0 \\
\hline
\end{tabular}
\label{tab:tabs4}
\end{table}

\subsubsection{PoseInj Module}
This fine-tuning stage aims to improve camera motion tracking through the explicit use of relative camera poses:
\[
\mathcal{L}' = \Big\| \epsilon_t - \mathcal{W}\big([z_t, \hat{l}^{\text{ego}}_1, E(P^{rel}, K)], \mathcal{V}^\text{exo}, t \big) \Big\|^2.
\]

To achieve this, we provide the training pipeline with an additional condition: the Plücker embedding of relative poses, adhering to the training configuration detailed in Table~\ref{tab:tabs4}. In this stage, we use LoRA to fine-tune only the self-attention modules. The LoRA components introduce 210M trainable parameters. For the inference stage, we follow the same configuration settings described in Table~\ref{tab:tabs2}.

\section{Impact of Inference Steps and Comparison to Exo2Ego-V}
{As mentioned in the introduction, we proposed our own baseline, VAWAN, because foundation models and smaller pretrained models optimize for different objectives: smaller models focus on minimizing distortion, while foundation models prioritize perceptual quality. Nevertheless, for completeness, as illustrated in Table~\ref{tab:inf_steps}, we also provide a comparison with Exo2Ego-V\cite{liu2024exocentric} for the \textit{Bike} subcategory, additionally highlighting the effects of different numbers of inference steps in our evaluation of Exo2EgoSyn.}

{We found that using fewer inference steps removes less Gaussian noise from the signal. As a result, each pixel tends to approximate the average of its neighborhood, which reduces MSE and increases PSNR, even though the visual quality is worse. For this reason, we did not report rows with artificially high PSNR in our main table, emphasizing perceptual quality over metric values. For the full comparison, see our main table and Table 1 in Exo2Ego-V~\cite{liu2024exocentric}.}

\begin{table}[h]
\caption{\textbf{Comparison with Exo2Ego-V~\cite{liu2024exocentric}.} Superscripts indicate the number of inference denoising steps used in the diffusion model.}
\centering
\begin{tabular}{lccc}
\toprule
Framework & \multicolumn{3}{c}{Bike} \\
\cmidrule(lr){2-4}
          & PSNR↑ & SSIM↑ & LPIPS↓ \\
\midrule
Exo2Ego-V  & 16.3101 & 0.4130 & 0.4861 \\
Ours$^{40}$ & 15.1319 & 0.3895 & 0.5052 \\
Ours$^{20}$ & 15.9468 & 0.4014 & 0.4995 \\
Ours$^{10}$ & 16.1219 & 0.4021 & 0.4982 \\

\bottomrule
\end{tabular}
\label{tab:inf_steps}

\end{table}
\section{User Study Materials}
\begin{figure}[h]
  \parbox{\linewidth}{
  \centering
    \includegraphics[width=0.5\linewidth]{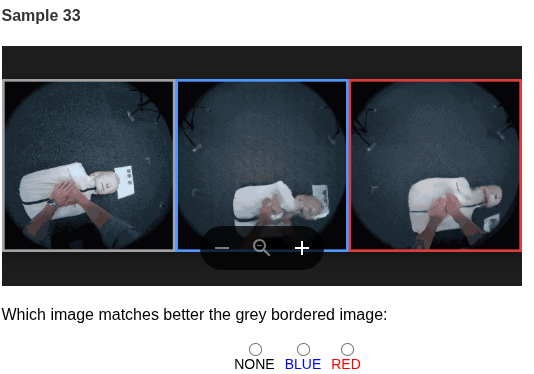}
    \caption{\textbf{User Study Sample.}}
    \label{fig:figs36}
  }
  \vspace{-0.15in}
\end{figure}

We provide additional details on our user study, which evaluates the proposed Exo2EgoSyn model in comparison to the baseline VAWAN. In Fig. \ref{fig:figs36} we show the interface for a
single sample given to the user. A clip is shown with two
parts, which the user should compare and rate. The order of
the samples is random. The instructions given to the users
at the start of the study are provided below:

\textit{Thank you for taking part in our user study! For each question you will see a row of 3 video frames. The reference video is outlined in \textcolor{darkgray}{grey}. The two candidate video frames are outlined in \textcolor{blue}{blue} and \textcolor{red}{red}.}

\textit{\textcolor{darkgray}{\underline{Grey}} Video Frames: Ground-truth reference}.

\textit{\textcolor{blue}{\underline{Blue}} Video Frames: Candidate that attempts to resemble the grey reference frames}.

\textit{\textcolor{red}{\underline{Red}} Video Frames: Candidate that attempts to resemble the grey reference frames}.

\textit{There are 100 samples in total. For each sample, decide which candidate most closely matches the content of the grey reference frames. Please base your decision on the accuracy of the \textbf{hand-over-object interaction, the visual quality}, and \textbf{the camera motion tracking}.}

\textit{Options:}

\textit{\textcolor{blue}{BLUE}: The blue video resembles the grey video better}

\textit{\textcolor{red}{RED}: The red video resembles the grey video better}

\textit{NONE: Both options equally well represent the grey}

\textit{After rating all sequences, click “Submit” to record your responses. Your feedback is invaluable—thank you!}

As outlined in Section 4.2, we conducted a user study with 10 participants who cast a total of 919 votes. The results show a clear preference for our model, which received 515 votes, compared to 215 votes for \textit{"None"} and 189 votes for the VAWAN baseline.
\section{Additional Qualitative Visualizations}
\begin{figure*}[p]
  \includegraphics[width=\linewidth]{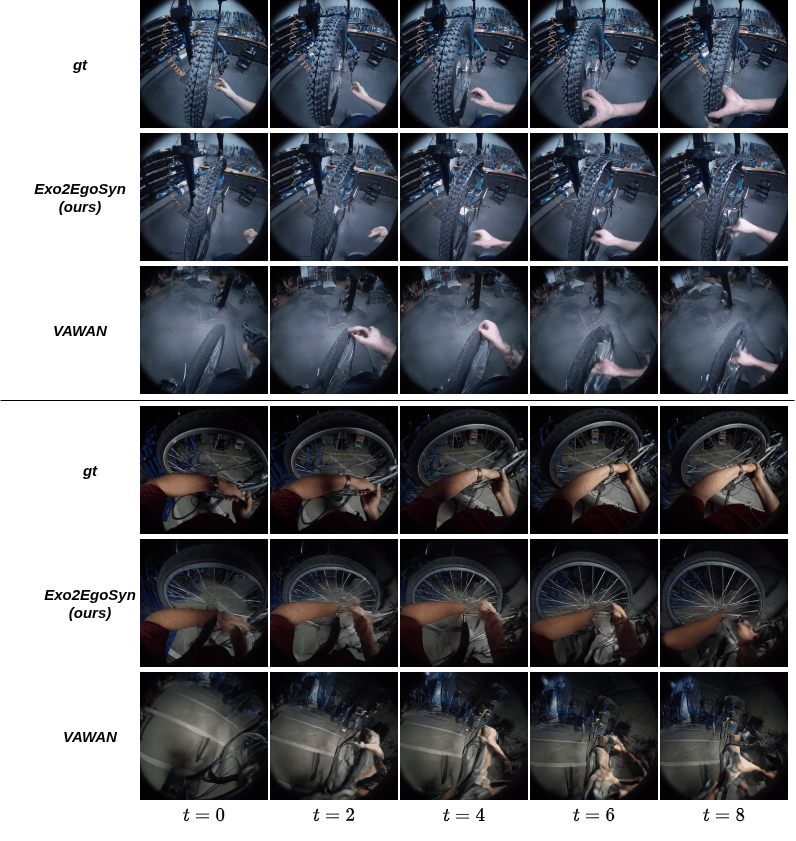}
  \caption{\textbf{Qualitative Results.} Bike.}
  \label{fig:figs31}
\end{figure*}

\begin{figure*}[t]
  \includegraphics[width=\linewidth]{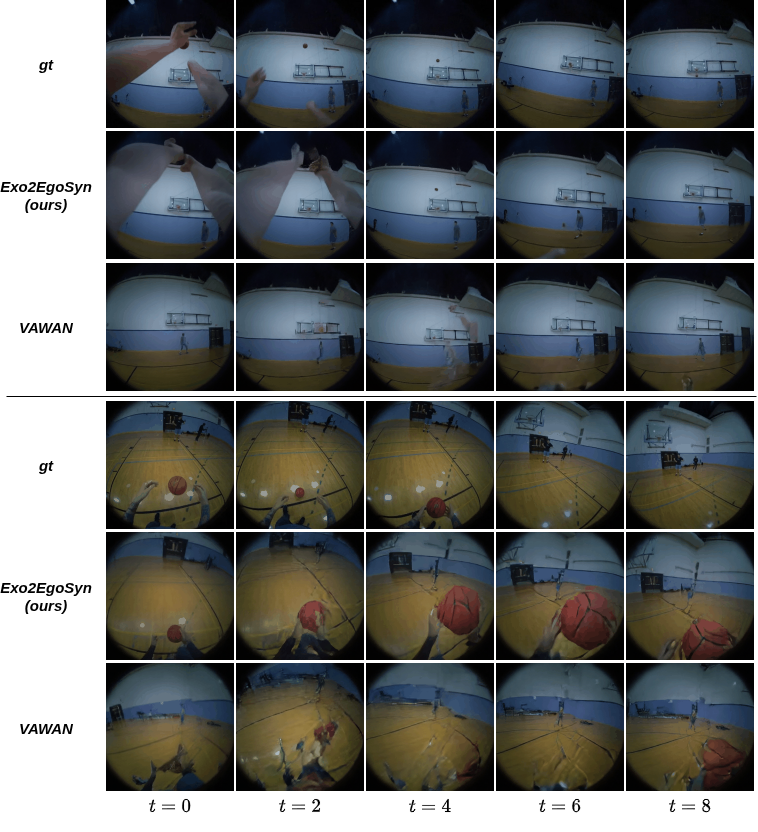}
  \caption{\textbf{Qualitative Results.} Basketball.}
  \label{fig:figs32}
\end{figure*}

\begin{figure*}[t]
  \includegraphics[width=\linewidth]{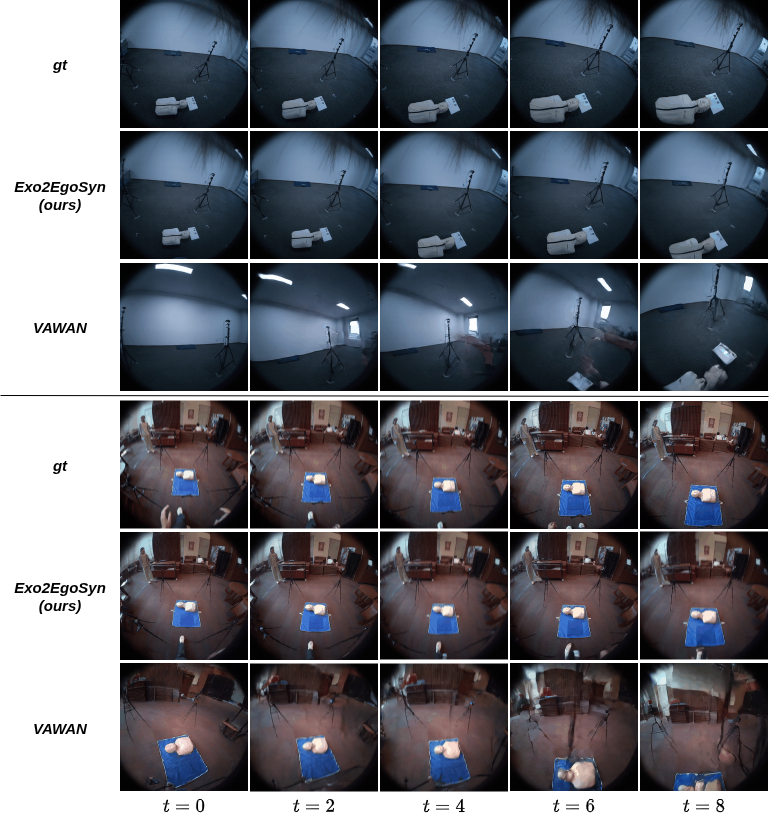}
  \caption{\textbf{Qualitative Results.} CPR.}
  \label{fig:figs33}
\end{figure*}

\begin{figure*}[t]
  \includegraphics[width=\linewidth]{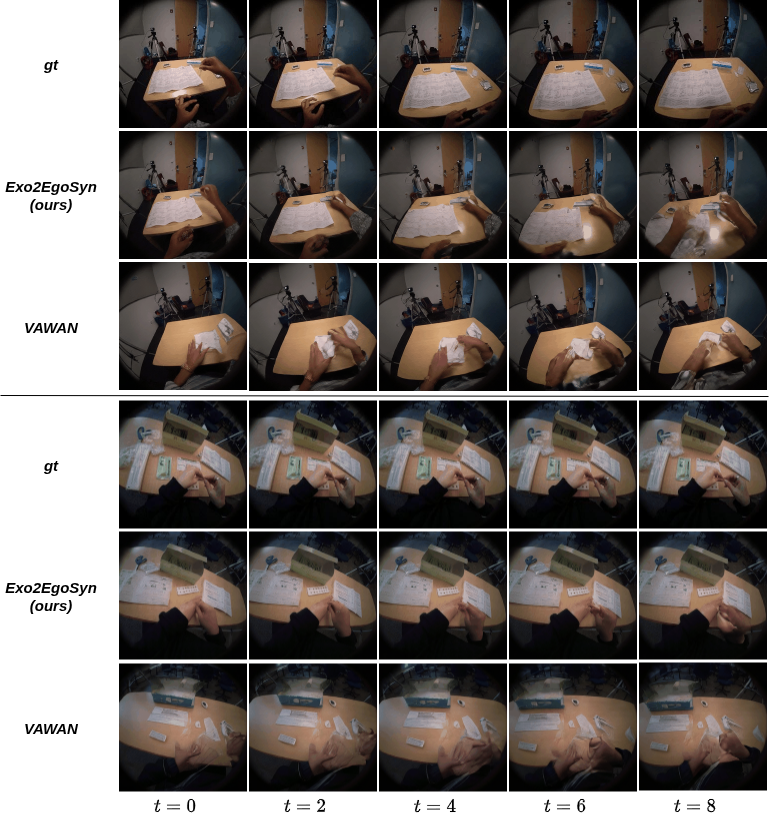}
  \caption{\textbf{Qualitative Results.} Covid Test.}
  \label{fig:figs34}
\end{figure*}

Extending the qualitative analysis in the paper, we present further visual comparisons for all categories in our evaluation. Fig. \ref{fig:figs31}, Fig. \ref{fig:figs32}, Fig. \ref{fig:figs33}, and Fig. \ref{fig:figs34} illustrate example outputs from the Bike, Basketball, CPR, and Covid Test categories, respectively, demonstrating the performance of our proposed model against the baseline.

\section{Analysis of Failure Modes}
\begin{figure*}[t]
  \includegraphics[width=\linewidth]{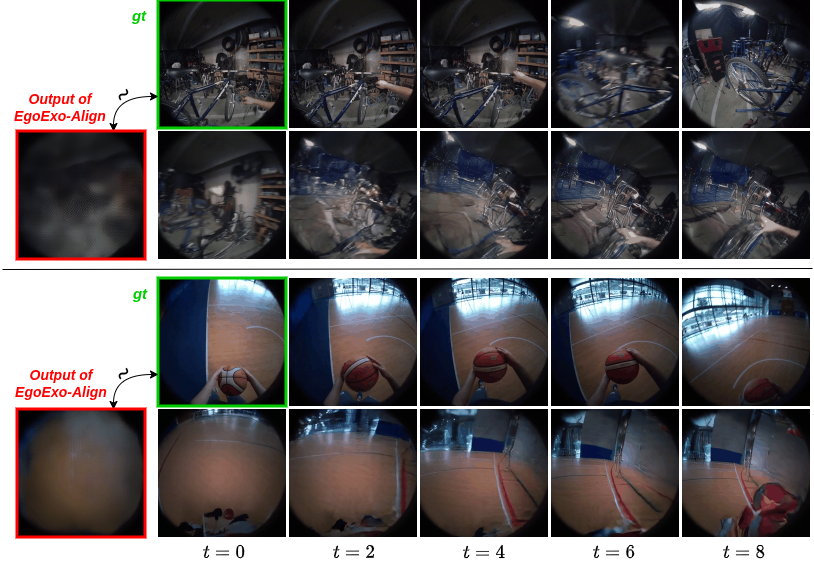}
  \caption{\textbf{Failure Cases.} Top: A case where the EgoExo-Align output (red) fails to accurately reconstruct the first frame of the ego video (green). This initial misalignment, combined with fast camera motion in the ground-truth sequence (see $t=6$), causes the model to produce poor results. Bottom: Here, EgoExo-Align produces a textureless conditioning frame. While the model successfully mimics the camera motion (compare predicted and ground-truth rows), it fails to generate the essential elements of the scene, namely the ball and the detailed basketball court.}
  \label{fig:figs35}
\end{figure*}

An analysis of our model's limitations reveals specific failure modes. Performance degrades primarily when the EgoExo-Align module produces a poor initial frame prediction, or when the predicted frame, while semantically correct, lacks informative visual content to guide the subsequent generation. For instance, an accurately predicted but textureless image of a basketball court floor provides insufficient cues for the model to generate a meaningful subsequent action. Failures also occur during actions with very fast motion, leading to blur in the egocentric view. Fig. \ref{fig:figs35} visualizes examples of these failure cases.

\section{Future Work}
Although prior work~\cite{liu2024exocentric} has begun exploring pretrained models (a 2D image generative U-Net adapted from Stable Diffusion \cite{blattmann2023stable}) for exocentric-to-egocentric generation, these methods are not built upon foundation-scale video generative models. Instead, they typically rely on a two-stage pipeline: a frame-to-frame generation model is trained first, and a motion module is later finetuned to stitch individual frames into a coherent sequence. In contrast, we are the first to build on a foundation video model and generate the full video frames at once, adhering to the principles and capabilities of modern large-scale video generation.

However, generating the full video at once introduces a notable challenge: precise frame-level alignment of camera motion. While our model reliably captures the overall camera trajectory (e.g., a left-to-right sweep), 
it may shift the exact frame where a directional change occurs, for example, executing a turn slightly before or after the ground-truth frame $t$.
This behavior is an inherent consequence of holistic sequence generation. Methods that operate frame-by-frame, followed by temporal stitching, naturally maintain strict per-frame motion cues, which our one-pass approach does not explicitly enforce.
Thus, a promising direction for future work is to augment our single-pass foundation-model pipeline with mechanisms that enable accurate, frame-specific adherence to camera pose transitions—without reverting to frame-by-frame prediction. Achieving such fine-grained temporal alignment while preserving the advantages of full-sequence generation represents an important next step toward robust and precise cross-view video synthesis.




\end{document}